# Results of improved fractional/integer order PDE-based binarization model

**Uche A. Nnolim***

*Department of Electronic Engineering, Faculty of Engineering, University of Nigeria, Nsukka, Enugu, Nigeria*

*uche.nnolim@unn.edu.ng*

**Abstract:** *In this report, we present and compare the results of an improved fractional and integer order partial differential equation (PDE)-based binarization scheme. The improved model incorporates a diffusion term in addition to the edge and binary source terms from the previous formulation. Furthermore, logarithmic local contrast edge normalization and combined isotropic and anisotropic edge detection enables simultaneous bleed-through elimination with faded text restoration for degraded document images. Comparisons of results with state-of-the-art PDE methods show improved and superior results.*

## 1. Introduction

In previous work, a simplified PDE-based text binarization model was proposed and developed with good results. However, it was noted that it lacked the capability of simultaneous faded text restoration and bleed-through elimination. Also, the scheme could not completely eliminate stains that had similar edge profiles with the desired text. Consequently, the removal of such defects led to loss of text information in certain regions of the images. Thus, in this study, an improved scheme is proposed to address such issues, while maintaining the simplified and effective framework of PDE-based binarization models [1].

The outline of the work starts with the section 2, which presents a brief overview of current and state-of-the-art PDE-based binarization models. The third section outlines the proposed scheme, while the fourth section involves results and comparisons; the final section concludes the study.

## 2. Overview of PDE-based binarization models

The PDE schemes include Cheriet [2], Nwogu et al [2], Mahani et al [2], Kumar et al [3], Bella et al [4], Guemri and Drira [5], Wang et al [6], Wang and He [7], Jacobs and Momoniat [8] [9]. Others are; Rivest-Henault et al [10], Drira and Yagoubi [2], Chen et al [11], Huang et al [12], Guo et al [2] [13], Guo and He [14], Zhang et al [15], Feng [16], Nnolim [1] [17] [18] and Du and He [19].

Prior to the work of Du and He, the typical PDE-based frameworks usually consisted of binarization source and diffusion terms. However, the previously proposed scheme utilized edge term and source terms [1], while Du and He's scheme replaces the source term with contrast enhancement terms. The proposed scheme utilizes fast pre-processing methods to complement the improved proposed model [17].

## 3. Proposed model

The modified scheme enables control of noise in addition to stain removal, while performing the binarization. Thus, the need for the three terms. Also, in order to further enhance the performance of the edge detection term, a modified local contrast scheme was used to enhance edges [17]. However, this is performed after reducing the magnitude of the stain component via linear and nonlinear attenuation configurations [17]. The proposed (integer order) PDE model was given as;

$$\frac{\partial u}{\partial t} = c_s U_s + c_e U_e + c_d U_d \tag{1}$$

The parameters include the binary source term ($U_s$) and its coefficient, $c_s$, the edge term ($U_e$) and its coefficient, $c_e$ and the diffusion term ($U_d$) and its coefficient, $c_d$. The coefficient of the source term is usually set to unity for document images with minimal to degradation. However, it can be increased for images with faded text with some stains, since the latter would be handled with the edge term coefficient. The diffusion term coefficient can be tuned based on the amount of noise in the document image and will also depend on the noisiness of the dataset used. The details and analysis of the functions and parameters for the proposed improved model can be found in [17]. The fractional version of the PDE is given as;

$$\frac{\partial^\alpha u}{\partial t^\alpha} = c_s U_s + c_e U^\alpha{}_e + c_d U^\alpha{}_d \tag{2}$$

The numerical implementations of both integer and fractional order were also developed [17]. The improved model adequately rectified the issue of simultaneous faded text restoration, noise and stain removal [17]. It also addressed the problems of stains with edge profiles similar to desired foreground text to some degree but had issues for cases with extreme bleed-through, noisy images with highly faint text and poor contrast.

## 4. Experiments and comparisons

We performed numerous experiments to evaluate the performance of the new proposed algorithm (NPA) using the DIBCO datasets (2009 – 2014, 2016-2017) [20]. Additionally, the effectiveness of the model was validated when compared with state-of-the-art PDE methods using the various objective measures from the literature [21]. These included models by Wang and He (WH) [7], Rivest-Hénault et al (RMC) [10], Jacobs and Momoniat (JM) [8], Wang et al (WYH) [6], FENG [16], Guo et al (GHZ) [2], Zhang et al (ZHG) [15], Guo et al (GHW) [13] and the previous proposed algorithm (PPA) [1]. Furthermore, the scheme was compared with the state-of-the-art deep learning techniques and showed competitive results for several of the methods [17] for a fraction of the structural, computational and time costs.

We present the results for varying parameter settings for the source, edge and diffusion terms in Figs. 1 to 4. We observe the degradations for the various document images and the effects of contributions of the individual terms in the binarization outcomes.

In Fig. 1, we keep the edge coefficient at unity, while increasing the source coefficient using nonlinear and linear stain attenuation configurations [17]. This enables the reduction of the effect of the stains, while boosting or preserving the edge strength of the text in the degraded document images. This is clearly observed in Fig. 1(e) as we increase the source contribution, enabling the preservation of text detail and almost completely eliminating the stamp stain.

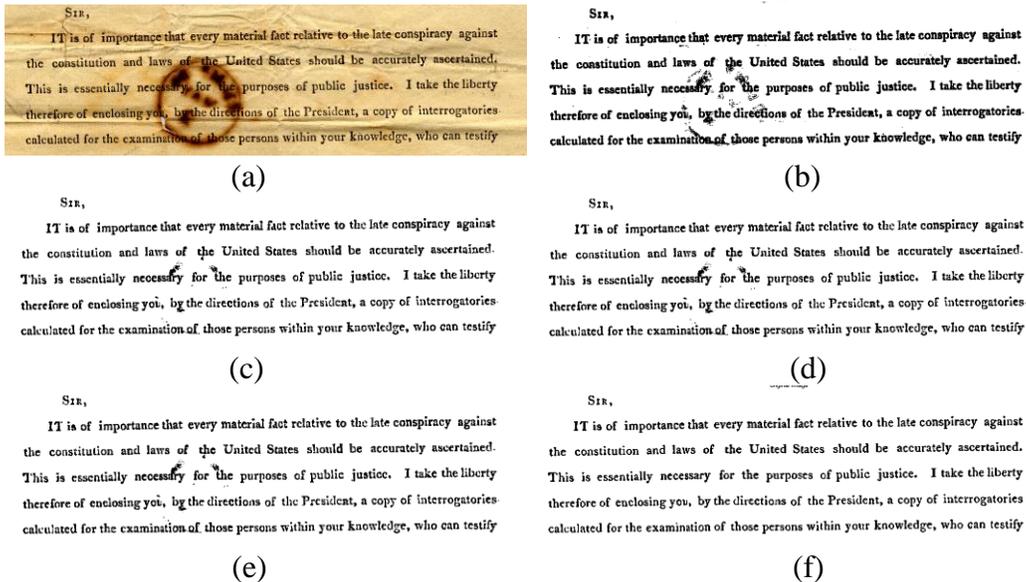

Fig. 1. (a) Original image processed with (b) NPA ($c_s = 2.5$; $c_e = 1$) using linear stain attenuation (c) NPA ($c_s = 2.5$; $c_e = 1$) using nonlinear stain attenuation (d) NPA ($c_s = 3$; $c_e = 1$) using nonlinear stain attenuation (e) NPA ($c_s = 4$; $c_e = 1$) using nonlinear stain attenuation (f) ground truth

In Fig. 2, we also determine the extent of the large oil stain in the image by varying the parameters using the stain attenuation pre-processing schemes. This prevents the loss of text details or fading text as we eliminate the stains, which was a weakness in the previous model. A similar effect is observed in Fig. 3, though the stain outline is still observed. Additional attempts to remove the outline would result in loss of text details for regions sharing similar intensities.

In Fig. 4, we observe that the NPA yields improved stain removal without fading text, unlike the previous proposed approach (PPA).

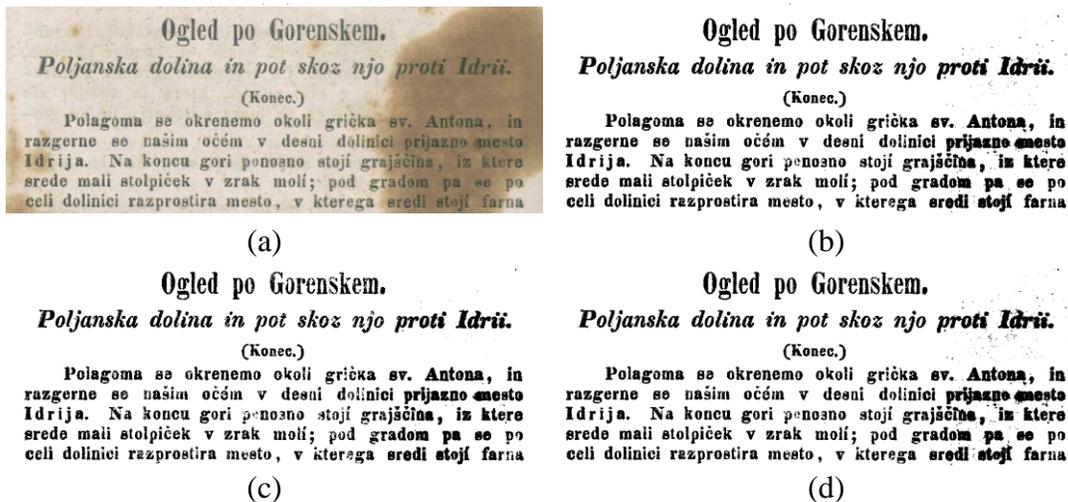

Fig. 2. (a) Original image processed with (b), (c) and (d) NPA (cs = 2 to 2.5; ce = 1)

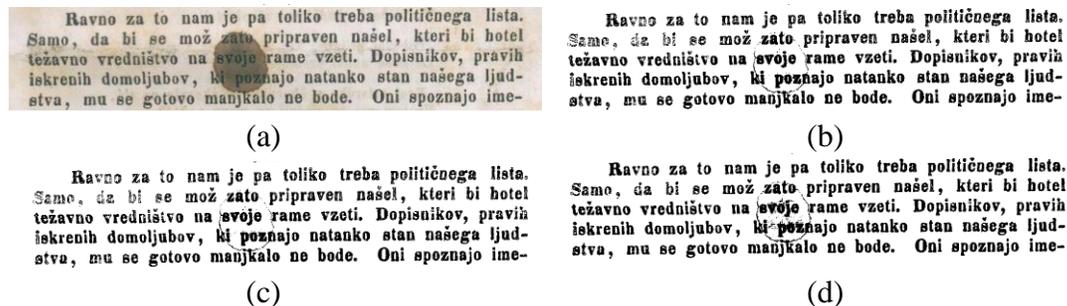

Fig. 3. (a) Original image processed with (b) NPA (c) NPA (d) (cs = 2 to 2.5; ce = 1)

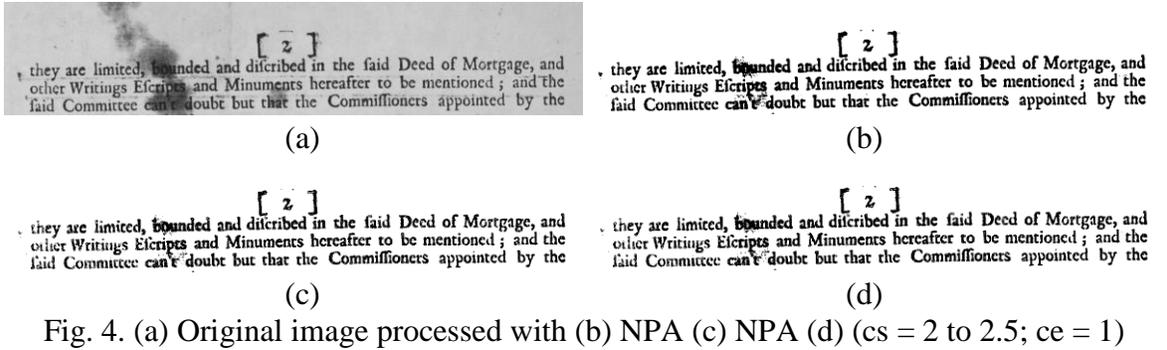

Fig. 4. (a) Original image processed with (b) NPA (c) NPA (d) (cs = 2 to 2.5; ce = 1)

We compare visual results of various PDE-based models for the sample DIBCO images in Fig. 5 and stain removal capability of the PPA and NPA are clear. Additionally, the improved model shows simultaneous stain removal, binary text enhancement and avoidance of faded text. Thus, the weaknesses of the other approaches with regard to stain mitigation is clear.

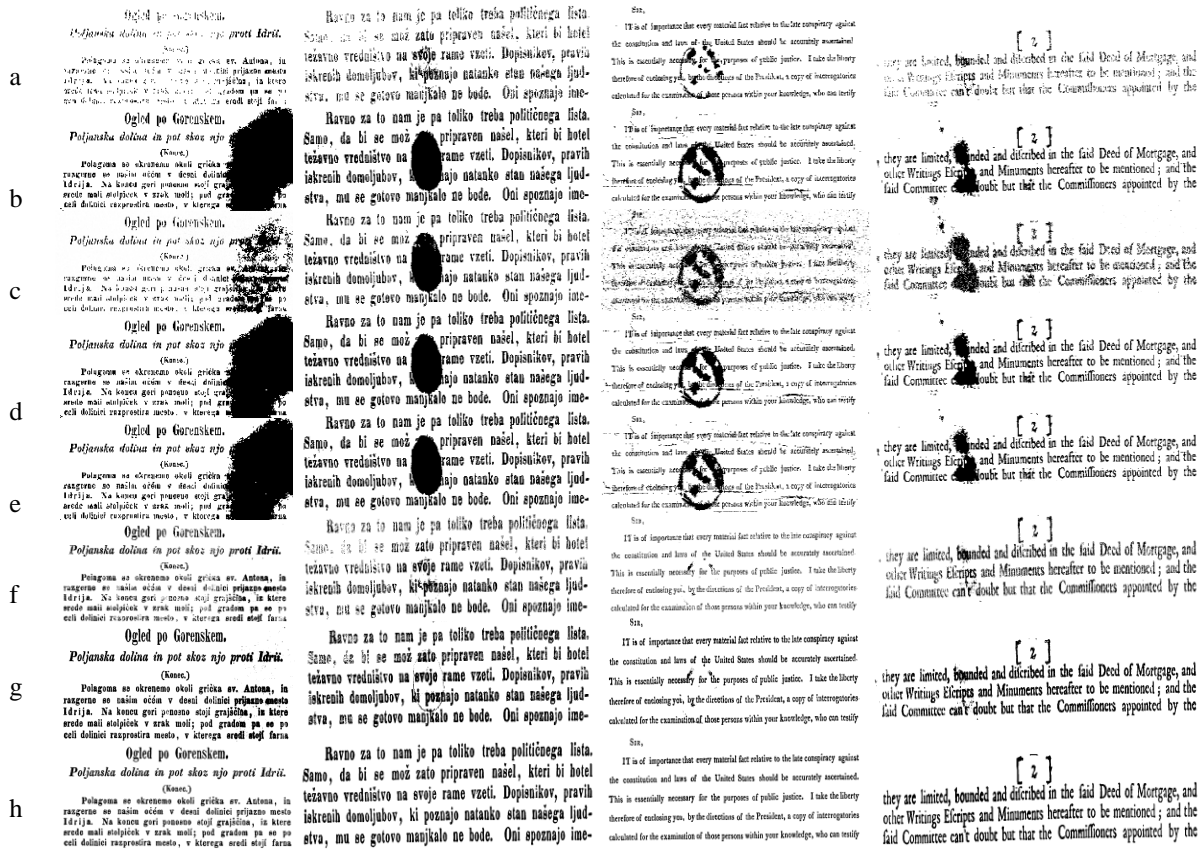

Fig. 5. Binarization results of various PDE models on stained document images (a) ZHG (b) GHZ (c) JM (d) FENG (e) GHW (f) PPA (g) NPA (h) Ground-truth

The numerical results obtained via relevant objective metrics [1] for DIBCO 2009 to 2014 and 2016 series are shown in Table 1 from [1] [17], indicating the superior performance of the NPA compared to the other PDE-based algorithms. Thus, the numerical values are consistent with the earlier observed visual results.

Table 1 Performance comparison of NPA with other PDE binarization algorithms

| Algorithms \ Metrics | FM (%) | $F_{PS}$ (%) | PSNR (dB) | DRD | NRM (%) |
|---|---|---|---|---|---|
| WH | 80.37 | 85.17 | 16.25 | 7.95 | - |
| RMC | 75.29 | 77.62 | 15.15 | 18.33 | - |
| JM | 77.21 | 79.03 | 15.60 | 7.14 | - |
| WYH | 86.75 | 89.11 | 17.53 | 4.71 | - |
| FENG | 75.00 | 80.55 | 15.87 | 9.68 | - |
| GHZ | 85.78 | 88.44 | 17.60 | 5.75 | - |
| ZHG | 85.75 | 89.27 | 17.72 | 5.54 | - |
| GHW | 86.34 | 89.71 | 17.68 | 6.25 | - |
| PPA | 88.00 | 90.30 | 18.11 | 3.99 | 0.07 |
| NPA | **89.03** | **91.31** | **18.78** | **3.47** | **0.06** |

## 5. Conclusion

The results and comparisons of a proposed improved PDE binarization model [17] confirm improvements in simultaneous stain removal and text enhancement for degraded document images. Simple pre-processing schemes to boost edge strength and intensity, while weakening stain profiles enhances the performance of the improved model. This resolves the issues of the previously developed PDE binarization model with the inclusion of diffusion term in addition to the source and edge terms. Future work will focus on improving on the model for document images with extreme bleed-through, stains with similar intensity profiles with foreground text and highly faded text in noisy backgrounds [18].

# References

[1] U. A. Nnolim, "Dynamic Selective Edge-Based Integer/Fractional Order Partial Differential Equation For Degraded Document Image Binarization," *International Journal of Image and Graphics,* vol. 21, no. 3, pp. 1-31, July 2021.

[2] J. Guo, C. He and X. Zhang, "Nonlinear edge-preserving diffusion with adaptive source for document images binarization," *Applied Mathematics and Computation,* vol. 351, pp. 8-22, 2019.

[3] S. S. Kumar, P. Rajendran, P. Prabaharan and K. Soman, "Text/Image region separation for document layout detection of old document images using non-linear diffusion and level set," in *6th International Conference on Advances in Computing and Communications (ICACC)*, 2016.

[4] F. Z. A. Bella, M. E. Rhabi, A. Hakim and A. Laghrib, "Reduction of the non-uniform illumination using nonlocal variational models for document image analysis," *J. Frankl. Inst.-Eng. Appl. Math,* vol. 355 , pp. 8225-8244, 2018 .

[5] K. Guemri and F. Drira, "Adaptative shock filter for image characters enhancement and denoising," in *6th International Conference of Soft Computing and Pattern Recognition (SoCPaR)*, 2014.

[6] Y. Wang, Q. Yuan and C. He, "Indirect diffusion-based level set evolution for image segmentation," *Appl. Math. Model.,* vol. 69 , p. 714–722, 2019.

[7] Y. Wang and C. He, "Binarization method based on evolution equations for document images produced by cameras," *Journal of Electronic Imaging,* vol. 21, no. 2, p. 023030 , 2012.

[8] B. A. Jacobs and E. Momoniat, "A novel approach to text binarization via a diffusion-based model," *Applied Mathematics and Computation,* vol. 225, pp. 446-460, 2013.

[9] B. A. Jacobs and E. Momoniat, "A locally adaptive, diffusion based text binarization technique," *Applied Mathematics and Computation,* vol. 269, pp. 464-472, 2015.

[10] D. Rivest-Hénault, R. Moghaddam and M. Cheriet, "A local linear level set method for the binarization of degraded historical document images," *Int. J. Doc. Anal. Recognit. ,* vol. 15 , no. 2, p. 101–124, 2012.

[11] B. Chen, S. Huang, Z. Liang, W. Chen and B. Pan, "A fractional order derivative based active contour model for inhomogeneous image segmentation," *Appl. Math. Model. ,* vol. 65, p. 120–136 , 2019.

[12] C. Huang and L. Zeng, "Level set evolution model for image segmentation based on variable exponent p-laplace equation," *Appl. Math. Model,* vol. 40 , no. 17–18, p. 7739–7750, 2016.

[13] J. Guo, C. He and Y. Wang, "Fourth order indirect diffusion coupled with shock filter and source for text binarization," *Signal Processing,* vol. 171 , no. 107478, pp. 1-13, January 2020.

[14] J. Guo and C. He, "Adaptive shock-diffusion model for restoration of degraded document images," *Applied Mathematical Modelling,* vol. 79, pp. 555-565, 2020.

[15] X. Zhang, C. He and J. Guo, "Selective diffusion involving reaction for binarization of bleed-through document images," *Applied Mathematical Modelling,* vol. 81, pp. 844-854, January 20 2020.

[16] S. Feng, "A novel variational model for noise robust document image binarization," *Neurocomputing,* vol. 325, pp. 288-302, October 6 2019.

[17] U. A. Nnolim, "Improved integer/fractional order partial differential equation-based thresholding," *Optik - International Journal for Light and Electron Optics,* vol. 229, no. 2, pp. 1-10, March 2021.

[18] U. A. Nnolim, "Enhancement of degraded document images via augmented fourth order partial differential equation and Total Variation-based illumination estimation," *OPTIK-International Journal for Light and Electron Optics,* vol. 249, p. 168262, 2022.

[19] Z. Du and C. He, "Nonlinear diffusion equation with selective source for binarization of degraded document images," *Applied Mathematical Modelling ,* vol. 99 , p. 243–259, July 2 2021.

[20] I. Pratikakis, K. Zagoris, G. Barlas and B. Gatos, "ICFHR 2016 handwritten document image binarization contest (h-DIBCO 2016)," in *Proceedings of the 15th International Conference on Frontiers in Handwriting Recognition*, Shenzhen, China, 2016.

[21] M. Sokolova and G. Lapalme, "A systematic analysis of performance measures for classification tasks," *Information Processing & Management ,* vol. 45, pp. 427-437, 2009.